\documentclass[runningheads]{llncs}
\usepackage{graphicx}
\usepackage{amsmath,amssymb} 
\usepackage{color}
\usepackage[width=122mm,left=12mm,paperwidth=146mm,height=193mm,top=12mm,paperheight=217mm]{geometry}
\begin{document}
\pagestyle{headings}
\mainmatter

\title{FishEyeRecNet: A Multi-Context Collaborative Deep Network for Fisheye Image Rectification} 

\titlerunning{Yin et al.}

\authorrunning{Yin et al.}

\author{Xiaoqing Yin\textsuperscript{1,2}, Xinchao Wang\textsuperscript{3,4}, Jun Yu\textsuperscript{5},\\ Maojun Zhang\textsuperscript{2}, Pascal Fua\textsuperscript{4}, Dacheng Tao\textsuperscript{1}}

\institute{\textsuperscript{1}University of Sydney, \textsuperscript{2}National University of Defense Technology, \\ \textsuperscript{3}Stevens Institute of Technology, \textsuperscript{4}\'Ecole Polytechnique F\'eéd\'erale de Lausanne,\\ \textsuperscript{5}Hangzhou Dianzi University}


\maketitle

\begin{figure*}[htbp]
	\setlength{\abovecaptionskip}{0pt}
	\centering
	\graphicspath{figure1.pdf/}
	\includegraphics [width=1\textwidth]{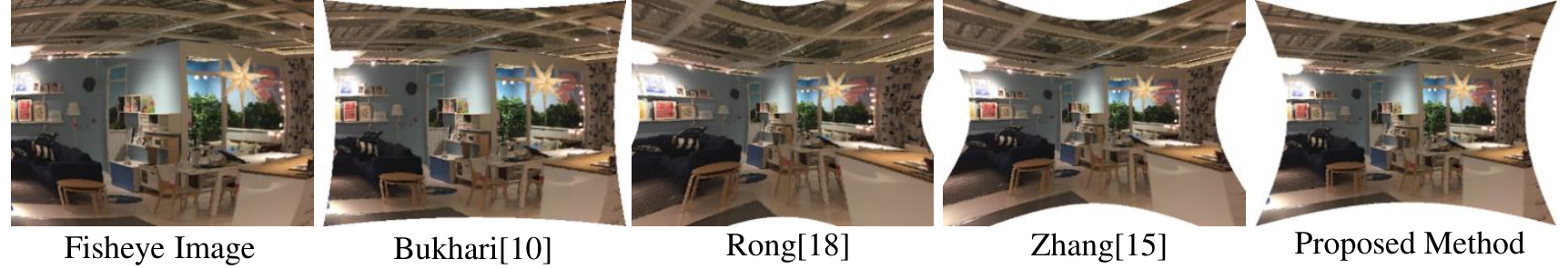}
	\caption{Our model performs rectification given a single fisheye image.}
	\label{fig1}
\end{figure*}

\begin{abstract}
Images captured by fisheye lenses violate the pinhole camera assumption and suffer from distortions. Rectification of fisheye images is therefore a crucial preprocessing step for many computer vision applications. In this paper, we propose an end-to-end multi-context collaborative deep network for removing distortions from single fisheye images. In contrast to conventional approaches, which focus on extracting hand-crafted features from input images, our method learns high-level semantics and low-level appearance features simultaneously to estimate the distortion parameters. To facilitate training, we construct a synthesized dataset that covers various scenes and distortion parameter settings. Experiments on both synthesized and real-world datasets show that the proposed model significantly outperforms current state of the art methods.
Our code and  synthesized dataset will be made publicly available.

\end{abstract}

\section{Introduction}

Fisheye cameras have been widely used in 
varieties of computer vision tasks, including virtual reality \cite{xiong1997creating,orlosky2014fisheye}, video surveillance \cite{drulea2014omnidirectional,decamp2010immersive},  automotive applications \cite{hughes2009wide,gehrig2005large} and depth estimation \cite{shah1994depth}, due to their large field of view.  
Images captured by such cameras however suffer from lens distortion,  and thus it is vital to perform rectification as a fundamental pre-processing step for subsequent tasks.
In recent years, active research work has been conducted on 
automatic rectification of fisheye images. In spite of the remarkable
progress, most existing rectification approaches  focus on handcrafted features \cite{sun2008calibration,mei2015radial,bukhari2013automatic,melo2013unsupervised,hughes2010equidistant,rosten2011camera,ying2004can,zhang2015line}, 
which have limited expressive power and  sometimes lead to unsatisfactory results.


We devise, to our best knowledge, the first end-to-end trainable deep convolutional neural network (CNN) for fisheye image rectification. Given a single fisheye image as input, our approach outputs the rectified image with distortions corrected, as shown in Fig. 1. Our method explicitly models the formation of fisheye images by first estimating the distortion parameters, during which step the semantic information is also incorporated. The warped images are then produced using the obtained parameters.  

We show the proposed model architecture in Fig. 2. We construct a deep CNN model to extract image features and feed the obtained features to a scene parsing network and a distortion parameter estimation network. The former network aims to learn a high-level semantic understanding of the scene, which is then provided to the latter network with the aim of boosting estimation performance. {The obtained distortion parameters, together with the input fisheye image and the corresponding scene parsing result, are then fed to a distortion rectification layer to produce the final rectified image and rectified scene parsing result.} 
The whole network is trained end-to-end. 

Our motivation for introducing the scene parsing network into the rectification model is 
that the learned high-level semantics can guide the distortion estimation.
{Previous methods usually rely on the assumption that straight lines in the 3D space have to 
	be straight after rectification. Nevertheless, 
	given an input image, it is difficult to determine which curved line should be straight in the 3D space. 
	The semantics could help to provide complementary information for this problem.
	For example, in the case of Fig.~5, semantic segmentation may potentially provide the knowledge that 
	the boundaries of skyscrapers should be straight after rectification but  those of the trees should not,
	and guide the rectification to produce plausible results shown in the last column of Fig.~5.
	Such high-level semantic supervision is, however, missing in the CNN used for extracting low-level features. 
	By incorporating the scene parsing branch, our model can therefore take advantage of both low-level features 
	and high-level semantics for the rectification process.
	
	To train the proposed deep network, we construct a synthesized dataset of visually high-quality images using the ADE20K~\cite{zhou2016semantic} dataset. {Our dataset consists of fisheye images and corresponding scene parsing labels, as well as rectified images and rectified scene parsing labels from ADE20K.} We synthesize both the fisheye images and the corresponding scene parsing labels. 
	{Samples are further augmented by adjusting distortion parameters to cover a higher diversity.
		
		We conduct extensive experiments to evaluate the proposed model on both the synthesized and real-world fisheye images. We compare our method with state of the art approaches on our synthesized dataset and also on real-world fisheye images using our model trained on the synthesized dataset. Our proposed model quantitatively and qualitatively outperforms state of the art methods and runs fast.
		
		Our contribution is therefore the first end-to-end deep learning approach for single fisheye image rectification. This is achieved by explicitly estimating the distortion parameters using the learned low-level features and under the guidance of high-level semantics. 
		Our model yields results superior to the current state of the art. More results are provided in the supplementary material. Our synthesized dataset and code will be made publicly available.

		\section{Related Work}
		
		We first briefly review existing fisheye image rectification and other distortion correction methods, and then discuss recent methods for low-level vision tasks with semantic guidance, which we also rely on in this work. 
		
		\subsection{Distortion Rectification}
		
		Previous work has focused on exploiting handcrafted features from distorted fisheye images for rectification. The most commonly used strategy is to utilize lines\cite{sun2008calibration,mei2015radial,bukhari2013automatic,melo2013unsupervised,hughes2010equidistant,rosten2011camera,ying2004can,zhang2015line,brand1993distorsions}, the most prevalent entity in man-made scenes, for the correction. The key idea is to recover the curvy lines caused by distortion to straight lines so that the pinhole camera model can be applied. 
		
		In the same vein, many methods follow the so-called plumb line assumption. Bukhari et al. \cite{bukhari2013automatic} proposed a method for radial lens distortion correction using an extended Hough transform of image lines with one radial distortion parameter. Melo et al. \cite{melo2013unsupervised}, on the other hand, used non-overlapping circular arcs for the radial estimation. However, in some cases especially for wide-angle lenses, these approaches yielded unsatisfactory results. Hughes et al. \cite{hughes2010equidistant} extracted vanishing points from distorted checkerboard images and estimated the image center and distortion parameters. This was, however, unsuitable for images of real-world scenes. 
		
		Rosten and Loveland \cite{rosten2011camera} proposed a method that transformed the edges of a distorted image to a 1-D angular Hough space and then optimized the distortion correction parameters by minimizing the entropy of the corresponding normalized histogram. The rectified results were, however, limited by hardware capacity. Ying et al. \cite{ying2004can} introduced a universal algorithm for correcting distortion in fisheye images. In this approach, distortion parameters were estimated using at least three conics extracted from the input fisheye image. {Brand~et~al.~\cite{brand1993distorsions} used a calibration grid to compute the distortion parameters. However, in many cases, it is difficult to obtain feature points whose world coordinates are known a priori}. Zhang et al. \cite{zhang2015line} proposed a multi-label energy optimization method to merge short circular arcs sharing the same or approximately the same circular parameters and selected long circular arcs for camera rectification. These approaches relied on line extractions in the first step, allowing errors to propagate to the final distortion estimation and compromise the results.

		The work most related to our method is \cite{rong2016radial}, where CNN was employed for radial lens distortion correction. However, the learning ability of this network was restricted to simulating a simple distortion model with only one parameter, which is not suitable for the more complex fisheye image distortion model. Moreover, this model only estimated the distortion model parameter and could not produce the final output in an end-to-end manner. 
		
		All the aforementioned approaches lack semantic information in the finer reconstruction level. Such semantics are, however, important cues for accurate rectification. By contrast, our model explicitly and jointly learns high-level semantics and low-level image features, and incorporates both streams of information in the fisheye image rectification process. The model directly outputs the rectified image and is trainable end-to-end.
		
		\subsection{Semantic Guidance}
		
		Semantic guidance has been widely adopted in low-level computer vision tasks. Liu et al. \cite{liu2017image} proposed a deep CNN solution for image denoising by integrating the modules of image denoising and high-level tasks like segmentation into a unified framework. Semantic information can thus flow into the optimization of the denoising network through a joint loss in the training process. Tsai et al. \cite{tsai2017deep} adopted a joint training scheme to capture both the image context and semantic cues for image harmonization. In their approach, semantic guidance was propagated to the image harmonization decoder, making the final harmonization results more realistic. Qu et al. \cite{qudeshadownet} introduced an end-to-end deep neural network with multi-context architecture for shadow removal from single images, where information from different sources were explored. In their model, one network was used to extract shadow features from a global view, while two complementary networks were used to generate features to obtain both the fine local details and semantic understanding of the input image, leading to state of the art performance.
		
		Inspired by these works, we propose to integrate semantic information to improve fisheye image rectification performance, which has, to our best knowledge, yet to be explored.

		\section{Methods}
		In this section, we describe our proposed model in detail. We start by providing a brief review of the fisheye camera model in Section 3.1, describe our network architecture in Section 3.2, and finally provide the definition of our loss function and training process in Section 3.3.
		
		\subsection{General Fisheye Camera Model}
		We start with the pinhole camera projection model, given as:
		\begin{equation}
			r = f\tan(\theta),
		\end{equation}
		where $\theta$ denotes the angle between the incoming ray and the optical axis, $f$ is the focal length, and $r$ is the distance between the image point and the principal point. 
		
		Unlike the pinhole perspective projection model, images captured by fisheye lenses follow varieties of projections, including stereographic, equidistance, equisolid and orthogonal projection \cite{hughes2010equidistant,kannala2004generic}.
		A general model is used for different types of fisheye lenses \cite{kannala2004generic}:
		\begin{equation}
			r(\theta)=k_1\theta+k_2\theta^3+k_3\theta^5+k_4\theta^7,	
		\end{equation}
		where $\left\{k_i\right\}(i=1,2,3,4)$ are the coefficients. Although Eq. (2) contains only four parameters, it is able to approximate all the projection models with high accuracy.
		
		
		Given pixel coordinates $(x,y)$ in the pinhole projection image, the corresponding image coordinates $(x',y')$ in the fisheye image can be computed: $x'=r(\theta)\cos(\varphi)$, $y'=r(\theta)\sin(\varphi)$, where $\varphi=\arctan((y-y_0)/(x-x_0))$, and $(x_0,y_0)$ are the coordinates of the principal point in the pinhole projection image.
		
		The image coordinates $(x',y')$ are then transformed to pixel coordinates $(xf,yf)$: $x_f=m_ux'+u_0$, $y_f=m_vy'+v_0$, 
		where $(u_0,v_0)$ are the coordinates of the principal point in the fisheye image, and $m_u,m_v$ denote the number of pixels per unit distance in the horizontal and vertical directions, respectively. We define $P_d=[k_1,k_2,k_3,k_4,m_u,m_v,u_0,v_0]$ as the parameters to be estimated, and describe the proposed model as follows.
		
		\subsection{Network Architecture}
		The proposed deep network is shown in Fig. 2. It aims to learn a mapping function from the input fisheye image to the rectified image in an end-to-end manner. Our basic idea is to exploit both the local image features and the contextual semantics 
		for the rectification process. To this end, we build our model by constructing a composite architecture consisting of four cooperative components as shown in Fig. 2: a base network (green box), a distortion parameter estimation network (gray box), a distortion rectification layer (red box) and a scene parsing network (yellow box).
		
		\begin{figure}[htb]
			\centering
			\includegraphics[width=1\textwidth]{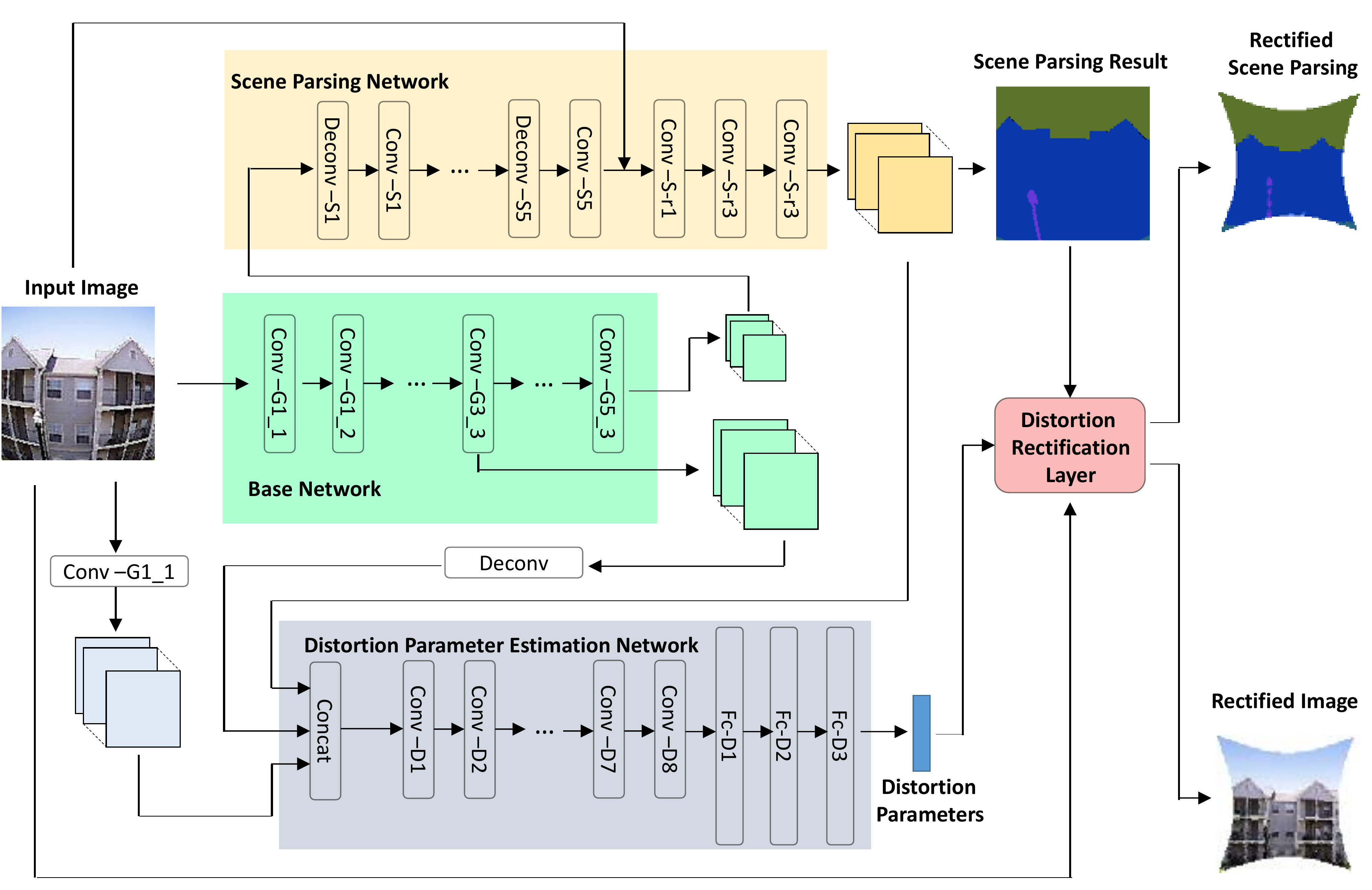}
			\caption{The overview of the proposed joint network architecture. This composite architecture consists of four cooperative components: a base network, a distortion estimation network, a distortion rectification layer, and a scene parsing network. The distortion parameter estimation network takes as input a concatenation of multiple feature maps from the base network and generates corresponding distortion parameters. Meanwhile, the scene parsing network extracts high-level semantic information to further improve the accuracy of distortion parameter estimation as well as rectification.
				The estimated parameters are then used by the distortion rectification layer to perform rectification on both the input fisheye image and the corresponding scene parsing results.}
			\label{fig2}
			\vspace{-0.3cm}
			
		\end{figure}
		
		In this unified network architecture, the base network is first used to extract low-level local features from the input image. The obtained features are then fed to the scene parsing network and the distortion parameter estimation network. The scene parsing network decodes the high-level semantic information to generate a scene parsing result for the input fisheye image. Next, the learned semantics are propagated to the distortion parameter estimation network to produce the estimated parameters. 
		{Finally, the estimated parameters, together with the input fisheye image and corresponding scene parsing result, are fed to the distortion rectification layer to generate the final rectified image and rectified scene parsing result. The whole network is trained end-to-end}. In what follows, we discuss each component in detail.

		\subsubsection{Base Network}
		
		The base network is built to extract both low- and high-level features for the subsequent fisheye image rectification and scene parsing tasks. Recent work suggests that CNNs trained with large amounts of data for image classification are generalizable to other tasks such as semantic segmentation and depth prediction. To this end, we adopt the VGG-net \cite{simonyan2014very} model for our base network, which is pre-trained on ImageNet for the object recognition task and fine-tuned under the supervision of semantic parsing and rectification.
		
		\subsubsection{Distortion Parameter Estimation Network}
		
		Our distortion parameter estimation network aims to estimate the distortion parameters $P_d$ discussed in section 3.1. This network takes as input a concatenation of multiple features maps: (1) The output of conv3-3 layer in the base network. Note that a deconvolution step is performed to raise the spatial resolution of feature maps; (2) The input image convolved with $3\times3$ learnable filters, which aims to preserve raw image information; and (3) The output of the scene parsing network. As shown in Section 4, we find that semantic priors help to eliminate the errors in distortion parameters. 
		
		In this distortion parameter estimation network, each convolutional layer is followed by a ReLU and a batch normalization \cite{ioffe2015batch}. We construct 8 convolutional layers with $3\times3$ learnable filters, where the number of filters is set as 64, 64, 128, 128, 256, 256, 512, and 512, respectively. Pooling layers with kernel size $2\times2$ and stride 2 are adopted after every two convolutions. Three fully-connected layers are added at the end of the network to produce the parameters, where each of the first two layers has 1024 units. To alleviate over-fitting, drop-out \cite{srivastava2014dropout} is adopted after the final convolutional layer with a drop probability of 0.5. 
		
		\subsubsection{Distortion Rectification Layer}
		The distortion rectification layer takes as input the estimated distortion parameters $P_d$,
		the fisheye image, as well as the scene parsing result.
		It computes the  corresponding pixel coordinates and generates the rectified image and the 
		rectified scene parsing result.
		This makes the network end-to-end trainable. Details of the distortion rectification layer are described as follows.
		
		In the forward propagation, given pixel location $(x,y)$ in the rectified image $I_r$, the corresponding coordinates $(x_f,y_f)$ in the input fisheye image $I_f$ are computed according to the aforementioned fisheye image model:
		\begin{equation}
			\begin{cases}
				x_f=x_0'+\frac{m_ux}{\sqrt{x^2+y^2}}(\theta+\sum_{i=1}^{4} k_i\theta^{2i+1})\\
				y_f=y_0'+\frac{m_vy}{\sqrt{x^2+y^2}}(\theta+\sum_{i=1}^{4} k_i\theta^{2i+1}).
			\end{cases}
		\end{equation}
		The pixel value of location $(x,y)$ in the rectified image is then obtained using the bilinear interpolation:
		\begin{equation}
			\begin{aligned}
				I_{x,y}^r=\overline{\omega_x}\overline{\omega_y}I_f(\left\lfloor x_f \right\rfloor,\left\lfloor y_f \right\rfloor) + 
				\omega_x\overline{\omega_y}I_f(\left\lceil x_f \right\rceil,\left\lfloor y_f \right\rfloor)\\
				+\overline{\omega_x}\omega_yI_f(\left\lfloor x_f \right\rfloor,\left\lceil y_f \right\rceil) + 
				\omega_x\omega_yI_f(\left\lceil x_f \right\rceil,\left\lceil y_f \right\rceil),
			\end{aligned}
		\end{equation}
		where the coefficients are computed as: $\omega_x=x_f-\left\lfloor x_f \right\rfloor $, $\omega_y=y_f-\left\lfloor y_f \right\rfloor $ and $\overline{\omega_x}=1-\omega_x$, $\overline{\omega_y}=1-\omega_y$.
		
		In the back propagation, we need to calculate the derivatives of rectified image with respect to the estimated distortion parameters as well as to the fisheye image. For each pixel $I_{x,y}^r$, derivatives with respect to input fisheye image are computed as follows:
		\begin{equation}
			\begin{aligned}
				\frac{\partial I_{x,y}^r}{\partial I_{x_f,y_f}^f}=\overline{\omega_x}\overline{\omega_y}\delta(\left\lfloor x \right\rfloor=x_f)\delta(\left\lfloor y \right\rfloor=y_f)\\
				+\omega_x\overline{\omega_y}\delta(\left\lceil x \right\rceil=x_f)\delta(\left\lfloor y \right\rfloor=y_f)\\
				+\overline{\omega_x}\omega_y\delta(\left\lfloor x \right\rfloor=x_f)\delta(\left\lceil y \right\rceil=y_f)\\
				+\omega_x\omega_y\delta(\left\lceil x \right\rceil=x_f)\delta(\left\lceil y \right\rceil=y_f),
			\end{aligned}
		\end{equation}
		where $\delta(s)=1$ if $s$ is true and 0 otherwise. Derivatives with respect to the estimated parameters $P_d$ are computed as follows: 
		\begin{equation}
			\frac{\partial I_{x,y}^r}{\partial P_i}=\frac{\partial I_{x,y}^r}{\partial x_f}\cdot\frac{\partial x_f}{\partial P_i}
			+\frac{\partial I_{x,y}^r}{\partial y_f}\cdot\frac{\partial y_f}{\partial P_i},
		\end{equation}
		where
		\begin{equation}
			\begin{aligned}
				\frac{\partial I_{x,y}^r}{\partial x_f}=-\overline{\omega_y}I_f(\left\lfloor x_f \right\rfloor,\left\lfloor y_f \right\rfloor) + 
				\overline{\omega_y}I_f(\left\lceil x_f \right\rceil,\left\lfloor y_f \right\rfloor)\\
				-\omega_yI_f(\left\lfloor x_f \right\rfloor,\left\lceil y_f \right\rceil) + 
				\omega_yI_f(\left\lceil x_f \right\rceil,\left\lceil y_f \right\rceil),
			\end{aligned}
		\end{equation}
		and $\partial x_f/\partial P_i$ is obtained according to:
		\begin{equation}
			\begin{cases}
				\frac{\partial x_f}{\partial k_i}=\frac{m_{x}x\theta^{2i+1}}{\sqrt{x^2+y^2}}(i=1,2,3,4)\\
				\frac{\partial x_f}{\partial m_j}=\frac{x(\theta+\sum_{i=1}^{4} k_i\theta^{2i+1})}{\sqrt{x^2+y^2}}(j=u,v)\\
				\frac{\partial x_f}{\partial u_0}=1.
			\end{cases}
		\end{equation}
		Similarly, we can calculate $\partial I_{x,y}^r/\partial y_f$ and $\partial y_f/\partial P_i$.  
		
		\subsubsection{Scene Parsing Network}
		
		The scene parsing network takes as input the learned local features and is provided with the scene parsing labels for training. Our motivation for introducing this network is that, in many tasks, semantic supervision may benefit low-vision tasks as discussed in Section 2.2. In our case, the scene parsing network outputs the semantic segmentations to provide high-level clues including the object contours in the image. Such segmentations provide much richer information compared to straight lines, which are treated as the only clue in many conventional distortion rectification methods. 
		
		In our implementation, we construct a decoder structure based on the outputs of VGG-Net. The decoder network consists of 5 convolution-deconvolution pairs with kernel size $3\times3$ for convolution layers and $2\times2$ for deconvolutions layers. The number of filters is set as 512, 256, 128, 64 and 32. As parts of the fisheye image are compressed due to distortion, the scene parsing results may lose some local details. We find that adding a refinement network can further improve the scene parsing accuracy. This refinement network takes the fisheye image and the initial scene parsing results as input and further refines the final results according to the details in the input image. Three convolutional layers are contained in the refinement network, with number of filters 32,32,16 and kernel size $3\times3$.
		
		As we will show in Section~4, in fact even without the scene parsing network, 
		our deep learning-based fisheye rectification approach already outperforms current state of the art approaches. 
		With the scene parsing network turned on, our semantic-aware rectification yields even higher accuracy.
		Since we feed to the network distorted segmentations as well as rectified ones, 
		our network can take advantage of such explicit segment-level supervision
		and potentially learn a segment-to-segment mapping,
		which helps to achieve better rectification.
		
		\subsection{Training Process}
		We aim to minimize the $L2$ reconstruction loss $L_r$ between the output rectified image $I_r$ and the ground truth image $I^{gt}$: 
		\begin{equation}
			L_r=\sum_x\sum_y\left\|I_{x,y}^{gt}-I_{x,y}^r\right\|_2^2.
		\end{equation}
		In addition to this rectification loss, we also adopt the loss $L_{sp}$ for the scene parsing task introduced by \cite{zhou2016semantic}. The final combined loss for the entire network is:
		\begin{equation}
			L=\lambda_1L_r+\lambda_2L_{sp},
		\end{equation}
		where $\lambda_1$ and $\lambda_2$ are the weights to balance the losses of fisheye image rectification and scene parsing, respectively. Thanks to the end-to-end trainable network architecture, our model can simultaneously learn the fisheye image rectification and the scene parsing tasks.
		
		In fact,  we also tried to train the network by minimizing errors directly on the distortion parameters. 
		However, we found that balancing distortion parameters of different natures, like $k_i$ and $(u_0,v_0)$, is very challenging. 
		We tested multiple strategies for balancing the terms in the distortion parameter loss, 
		and the best results we achieved are considerably lower than 
		those obtained using image reconstruction loss.
		
		We implement our model in Caffe \cite{jia2014caffe} and apply the adaptive gradient algorithm (ADAGRAD) \cite{duchi2011adaptive} to optimize the entire network. We set the initial learning rate as 1e-3 and reduce it by a factor of 10 every 200K iterations. In the joint training process, we start with training data from the ADE20K dataset to obtain an initial solution for both the fisheye image rectification and scene parsing tasks. During this initial training process, we set $\lambda_1=1$ and $\lambda_2=15$. Next, we fix the scene parsing part with $\lambda_2=0$, and fine-tune the rest of the network to achieve an optimal solution for fisheye image rectification. Note that, during this fine-tuning step, the scene parsing module propagates learned semantic information to the distortion parameter estimation network.
		
		By integrating the scene parsing model, the proposed network learns high-level contextual semantics like boundary features and semantic category layout and provides this knowledge to the distortion parameter estimation. For example, our network can produce straight-line contours for images with buildings and vehicles, which provides crucial clues for the rectification process.
		
		\section{Experiments}
		In this section, we discuss our experimental setup and results. We first introduce our data generation strategies in Section 4.1 and then compare our rectification results with those of the state of the art methods quantitatively in Section 4.2 and qualitatively in Section 4.3. We further show some scene parsing results in Section 4.4 and compare the runtime of our method and others in Section 4.5. We provide more results in the supplementary material.
		
		\subsection{Data Generation} 
		To train the proposed deep network for fisheye image rectification, we must first build a large-scale dataset. Each training sample should consist of a fisheye image, a rectification ground truth, and the scene parsing labels. To this end, we select a subset of the ADE20K dataset \cite{zhou2016semantic} with scene parsing labels and then follow the fisheye image model in Section 3.1 to create both the fisheye images and the corresponding scene parsing labels. During training, training samples are further augmented by randomly adjusting distortion parameters. The proposed dataset thus covers various scenes and distortion parameter settings, providing a wide range of diversities that potentially prevent over-fitting.
		
		Our training dataset includes 2,450 unique source images, each of which is used to generate 10 samples with various distortion parameter settings. Our test dataset contains 100 source images and 1000 samples generated using a similar strategy. We will make our dataset publicly available.

		\subsection{Quantitative Evaluation} 
		The dataset we constructed enables us to quantitatively assess our method. We run the proposed model and the state of the art ones on our dataset and evaluate them using standard metrics including PSNR and SSIM.
		{All the baseline models were realized according to the implementation details provided in corresponding papers. 
			The model \cite{rong2016radial} was trained on our simulated dataset, as done for ours.}

		{We show the quantitative comparisons in Tab.~1. Our method significantly outperforms existing methods in terms of both PSNR and SSIM. To further verify the semantic guidance, we add two experiments for the proposed method: 1) removing both the scene parsing network and the semantic loss, denoted as ``Proposed method - SPN - SL'', and 2) removing the semantic loss, but keeping the scene parsing network, denoted as ``Proposed method - SL''. 
			The networks are trained using the same settings. 
			The results indicate that the explicit semantic supervision does play an important role. 
			Robust feature extraction and semantic guidance contribute to more accurate rectification results.}
		
		\begin{table}
			\renewcommand{\arraystretch}{1.3}
			\newcommand{\tabincell}[2]{\begin{tabular}{@{}#1@{}}#2\end{tabular}}
			\caption{PSNR and SSIM scores of different algorithms on our test dataset.}
			\label{table_example}
			\centering
			\begin{tabular}{|c|c|c|}
				\hline
				{\bf Methods} & {\bf PSNR} & {\bf SSIM} \\
				\hline
				Bukhari \cite{bukhari2013automatic} & 11.47 & 0.2429\\
				\hline
				Rong \cite{rong2016radial} & 13.08 & 0.3356\\
				\hline 
				Zhang \cite{zhang2015line} & 12.52 & 0.2972\\
				\hline
				\tabincell{c}{Proposed method -SPN -SL} 
				& 14.39 & 0.3903\\
				\hline
				\tabincell{c}{Proposed method - SL} 
				& 14.47 & 0.3932\\
				\hline		
				Proposed method & \bf{14.96} & \bf{0.4129}\\
				\hline
			\end{tabular}
		\end{table}
		
		\begin{figure}[!htb]
			\centering
			\includegraphics [width=1\textwidth]{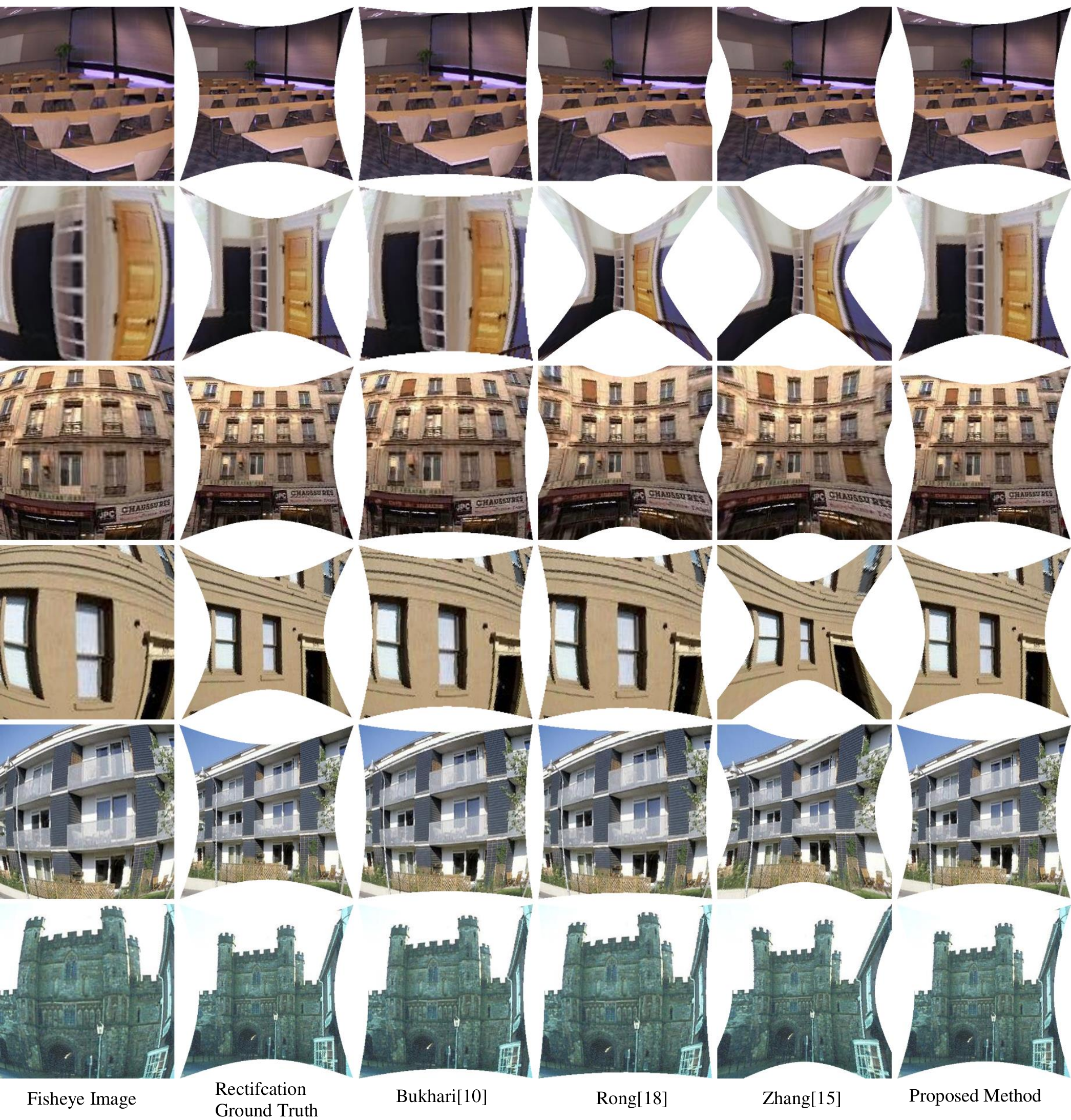}
			\vspace{-0.75cm}
			\caption{Qualitative results on our synthesized datasets. From left to right, we show the input, the ground truth, results of three state of the art methods (\cite{bukhari2013automatic,zhang2015line,rong2016radial}), and the result of our proposed approach. Our method achieves the best overall visual quality of all the compared methods.}
			\vspace{-0.5cm}
		\end{figure}
		
		\subsection{Qualitative Evaluation} 
		
		The qualitative rectification results on our synthesized dataset obtained by our method and the others are shown in Fig. 3. Our method produces results that are overall the most visually plausible and most similar to the ground truths, as evidenced by the fact that we restore the curvy lines to straight, which the other methods fail to do well. 
		
		\begin{figure}[htb]
			\setlength{\abovecaptionskip}{0pt}
			\centering
			\graphicspath{figure4.pdf/}
			\includegraphics [width=1\textwidth]{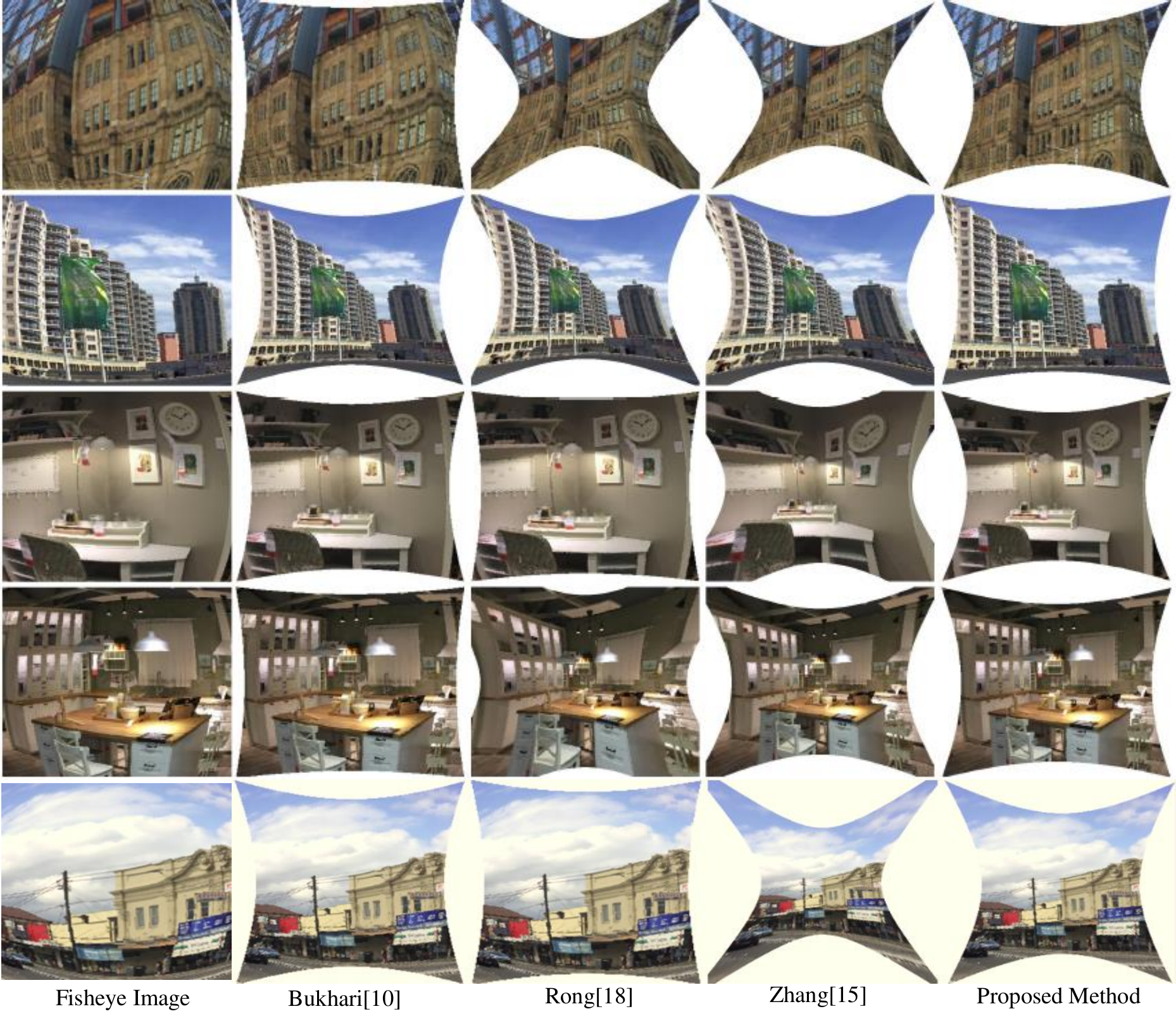}
			\caption{Qualitative results on real fisheye images. From left to right: the input image, results of three state of the art methods (\cite{bukhari2013automatic,zhang2015line,rong2016radial}), and results using our proposed method.}
			\vspace{-0.5cm}
		\end{figure}
		
		To show the effectiveness of our method on real fisheye images, we examine a test set of 650 real fisheye images captured using multiple fisheye cameras with different distortion parameter settings. {Samples of different projection types are collected, including stereographic, equidistance, equisolid angle and orthogonal~\cite{kannala2004generic}.} To cover a wide variety of scenarios, we collect samples from various indoor and outdoor scenes. Selective comparative results are shown in Fig. 4. Our method achieves the most promising visual  performance, which indicates our model trained on simululated dataset generalize well to real fisheye images.
		
		The results of \cite{bukhari2013automatic,zhang2015line} are fragile to
		the hand-crafted feature extraction. 
		In addition, the rectification of~\cite{zhang2015line} is very sensitive to the initial 
		value provided for the Levenberg-Marquardt~(LM) iteration process, 
		making it difficult to be deployed in real-world applications.
		The approach of \cite{rong2016radial}, on the other hand, is limited to a simple distortion model with  one parameter only,
		and thus it often fails to deal with more complex fisheye image distortion model with multiple types of paramters. 
		Our method, by contrast, is a fully end-to-end trainable approach for fisheye image rectification that learns 
		robust features under the guidance of semantic supervision.
		
		The results from both the synthesized dataset and the real fisheye dataset validate the effectiveness of our model, which uses synthesized data to learn how to perform fisheye image rectification given corresponding ground truth-rectified images. Our network learns both low-level local and high-level semantic features for rectification, which is, to our best knowledge, the first attempt at fisheye distortion rectification.

		\subsection{Scene Parsing Results}
		As shown in Table 1, even without the scene parsing module, our method already outperforms the other methods. With the guidance of the semantics, our method yields even better results in terms of PSNR and SSIM as shown in Table 1. To provide more insights into the scene parsing module, we show the parsing results obtained by the network in Fig. 5. It can be seen that the obtained parsing results are visually plausible, indicating that the network can produce semantic segmentation on distorted images, which may be further utilized by the rectification that takes place at a later stage. 
		
		We further show in Fig.~5 the rectified images produced by our model without and with the semantic guidance. The results without semantics are generated by removing the scene parsing network from the entire architecture. Our model without semantics, in spite of its already superior performance to other state of the art methods,
		still produces erroneous results like the distorted boundaries of the skyscraper and the vehicle shown in Fig. 5. 
		With the help of explicit semantic supervision, 
		our final model can potentially learn a segment-to-segment mapping for each semantic category,
		like the skyscraper or the car, and better guide the rectification  
		during testing.
		
		
		Regarding the influence of segmentation quality, 
		despite that we indeed observe some erroneous parsings in our experiments, 
		the imperfect segmentation results do help improve rectification for over 90\% of the cases.
		We expect the improvement to be even more significant with better segmentations. 
		As for the model generalization, since 
		the ADE20K benchmark~\cite{zhou2016semantic} comprises  objects of 150 classes
		and covers most semantic categories in daily life, 
		our model is able to handle most common objects.
		Handling unseen classes is left for further work.
		
		\subsection{Runtime}
		The run times of our methods and others are compared in Tab. 2. The methods of \cite{bukhari2013automatic,zhang2015line} rely on a minimazing complex objective function and time-consuming iterative optimization. Therefore, these approaches are difficult to accelerate by hardware-based parallelization and require much longer processing time on a $256\times256$ test image. On the contrary, our method can benefit from non-iterative forward process implemented on GPU. For example, when running the experiments on an Intel i5-4200U CPU, methods of \cite{bukhari2013automatic,zhang2015line} take over 60 seconds to generate one rectified result. Although our model is slower than \cite{rong2016radial}, the rectification performance is much better in terms of PSNR and SSIM.
		
		\begin{figure}[!htb] 
			\setlength{\abovecaptionskip}{0pt}
			\centering
			\graphicspath{figure5.pdf/}
			\includegraphics [width=1\textwidth]{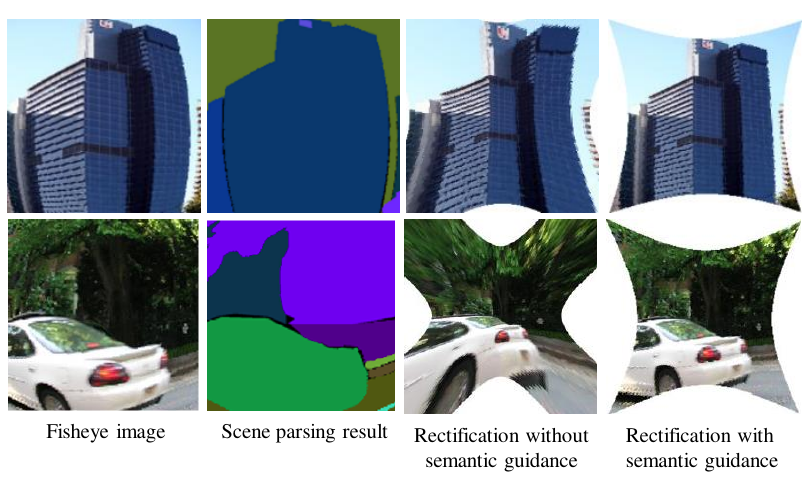}
			\vspace{-0.0cm}
			\caption{Qualitative rectification results obtained by our model without and with semantic guidance. With semantic supervision, the model can correct distortions that are otherwise neglected by the one without semantics. For example, the straight boundaries of the skyscraper and the shape of the vehicle can be better recovered.}
			\vspace{-0.0cm}
		\end{figure}
		
		
		\begin{table}
			\renewcommand{\arraystretch}{1.3}
			\caption{Run times of different algorithms on our test dataset.}
			\label{table_example}
			\centering
			\begin{tabular}{|c|c|}
				\hline
				{\bf Methods} & {\bf Average run time (seconds)}\\
				\hline
				Bukhari \cite{bukhari2013automatic} & 62.53 (Intel i5-4200U CPU)\\
				\hline
				Zhang \cite{zhang2015line} & 80.07 (Intel i5-4200U CPU)\\
				\hline
				Rong \cite{rong2016radial} & 0.87 (K80 GPU) \\
				\hline	
				Proposed method	& 1.31 (K80 GPU) \\ 	
				\hline
			\end{tabular}
		\end{table}

		\section{Conclusions}
		
		We devise a multi-context collaborative deep network for single fisheye image rectification. Unlike existing methods that mainly focus on extracting hand-crafted features from the input distorted images, which  have limited expressive power and are often unreliable, our method learns both high-level semantic and low-level appearance information for distortion parameter estimation. 
		Our network  consists of three collaborative sub-networks and is end-to-end trainable.
		A distortion rectification layer is designed to perform rectification on both the input fisheye image and corresponding scene parsing results. 
		For training, we construct a synthesized dataset covering a wide range of scenes and distortion parameters. 
		We demonstrate that our approach outperforms state of the art models on the synthesized and real fisheye images, both qualitatively and quantitatively. 
		
		In our further work, we will extend this framework to handle other distortion correction 
		tasks like the general radial lens distortion correction. 
		Also, we will explore to handle unseen semantic classes for rectification.

\bibliographystyle{unsrt}
\bibliography{egbib_arxiv}

\begin{thebibliography}{10}

\bibitem{xiong1997creating}
Yalin Xiong and Kenneth Turkowski.
\newblock Creating image-based vr using a self-calibrating fisheye lens.
\newblock In {\em Computer Vision and Pattern Recognition, 1997. Proceedings.,
  1997 IEEE Computer Society Conference on}, pages 237--243. IEEE, 1997.

\bibitem{orlosky2014fisheye}
Jason Orlosky, Qifan Wu, Kiyoshi Kiyokawa, Haruo Takemura, and Christian
  Nitschke.
\newblock Fisheye vision: peripheral spatial compression for improved field of
  view in head mounted displays.
\newblock In {\em Proceedings of the 2nd ACM symposium on Spatial user
  interaction}, pages 54--61. ACM, 2014.

\bibitem{drulea2014omnidirectional}
Marius Drulea, Istvan Szakats, Andrei Vatavu, and Sergiu Nedevschi.
\newblock Omnidirectional stereo vision using fisheye lenses.
\newblock In {\em Intelligent Computer Communication and Processing (ICCP),
  2014 IEEE International Conference on}, pages 251--258. IEEE, 2014.

\bibitem{decamp2010immersive}
Philip DeCamp, George Shaw, Rony Kubat, and Deb Roy.
\newblock An immersive system for browsing and visualizing surveillance video.
\newblock In {\em Proceedings of the 18th ACM international conference on
  Multimedia}, pages 371--380. ACM, 2010.

\bibitem{hughes2009wide}
C~Hughes, M~Glavin, E~Jones, and P~Denny.
\newblock Wide-angle camera technology for automotive applications: a review.
\newblock {\em IET Intelligent Transport Systems}, 3(1):19--31, 2009.

\bibitem{gehrig2005large}
Stefan~K Gehrig.
\newblock Large-field-of-view stereo for automotive applications.
\newblock In {\em Proc. of Workshop on Omnidirectional Vision, Camera Networks
  and Nonclassical cameras (OMNIVIS2005)}, 2005.

\bibitem{shah1994depth}
Shishir Shah and JK~Aggarwal.
\newblock Depth estimation using stereo fish-eye lenses.
\newblock In {\em Image Processing, 1994. Proceedings. ICIP-94., IEEE
  International Conference}, volume~2, pages 740--744. IEEE, 1994.

\bibitem{sun2008calibration}
Jie Sun and Jinhui Zhu.
\newblock Calibration and correction for omnidirectional image with a fisheye
  lens.
\newblock In {\em Natural Computation, 2008. ICNC'08. Fourth International
  Conference on}, volume~6, pages 133--137. IEEE, 2008.

\bibitem{mei2015radial}
Xiang Mei, Sen Yang, Jiangpeng Rong, Xianghua Ying, Shiyao Huang, and Hongbin
  Zha.
\newblock Radial lens distortion correction using cascaded one-parameter
  division model.
\newblock In {\em Image Processing (ICIP), 2015 IEEE International Conference
  on}, pages 3615--3619. IEEE, 2015.

\bibitem{bukhari2013automatic}
Faisal Bukhari and Matthew~N Dailey.
\newblock Automatic radial distortion estimation from a single image.
\newblock {\em Journal of mathematical imaging and vision}, 45(1):31--45, 2013.

\bibitem{melo2013unsupervised}
Rui Melo, Michel Antunes, Jo{\~a}o~Pedro Barreto, Gabriel Falcao, and Nuno
  Goncalves.
\newblock Unsupervised intrinsic calibration from a single frame using a.
\newblock In {\em Proceedings of the IEEE International Conference on Computer
  Vision}, pages 537--544, 2013.

\bibitem{hughes2010equidistant}
Ciaran Hughes, Patrick Denny, Martin Glavin, and Edward Jones.
\newblock Equidistant fish-eye calibration and rectification by vanishing point
  extraction.
\newblock {\em IEEE Transactions on Pattern Analysis and Machine Intelligence},
  32(12):2289--2296, 2010.

\bibitem{rosten2011camera}
Edward Rosten and Rohan Loveland.
\newblock Camera distortion self-calibration using the plumb-line constraint
  and minimal hough entropy.
\newblock {\em Machine Vision and Applications}, 22(1):77--85, 2011.

\bibitem{ying2004can}
Xianghua Ying and Zhanyi Hu.
\newblock Can we consider central catadioptric cameras and fisheye cameras
  within a unified imaging model.
\newblock {\em Computer Vision-ECCV 2004}, pages 442--455, 2004.

\bibitem{zhang2015line}
Mi~Zhang, Jian Yao, Menghan Xia, Kai Li, Yi~Zhang, and Yaping Liu.
\newblock Line-based multi-label energy optimization for fisheye image
  rectification and calibration.
\newblock In {\em Proceedings of the IEEE Conference on Computer Vision and
  Pattern Recognition}, pages 4137--4145, 2015.

\bibitem{zhou2016semantic}
Bolei Zhou, Hang Zhao, Xavier Puig, Sanja Fidler, Adela Barriuso, and Antonio
  Torralba.
\newblock Semantic understanding of scenes through the ade20k dataset.
\newblock {\em arXiv preprint arXiv:1608.05442}, 2016.

\bibitem{brand1993distorsions}
P~Brand, R~Mohr, and P~Bobet.
\newblock Distorsions optiques: correction dans un mod le projectif.
\newblock {\em 9dine congr\~{} s AFCET RFIA}, pages 87--98, 1993.

\bibitem{rong2016radial}
Jiangpeng Rong, Shiyao Huang, Zeyu Shang, and Xianghua Ying.
\newblock Radial lens distortion correction using convolutional neural networks
  trained with synthesized images.
\newblock In {\em Asian Conference on Computer Vision}, pages 35--49. Springer,
  2016.

\bibitem{liu2017image}
Ding Liu, Bihan Wen, Xianming Liu, and Thomas~S Huang.
\newblock When image denoising meets high-level vision tasks: A deep learning
  approach.
\newblock {\em arXiv preprint arXiv:1706.04284}, 2017.

\bibitem{tsai2017deep}
Yi-Hsuan Tsai, Xiaohui Shen, Zhe Lin, Kalyan Sunkavalli, Xin Lu, and Ming-Hsuan
  Yang.
\newblock Deep image harmonization.
\newblock {\em arXiv preprint arXiv:1703.00069}, 2017.

\bibitem{qudeshadownet}
Liangqiong Qu, Jiandong Tian, Shengfeng He, Yandong Tang, and Rynson~WH Lau.
\newblock Deshadownet: A multi-context embedding deep network for shadow
  removal.

\bibitem{kannala2004generic}
Juho Kannala and Sami Brandt.
\newblock A generic camera calibration method for fish-eye lenses.
\newblock In {\em Pattern Recognition, 2004. ICPR 2004. Proceedings of the 17th
  International Conference on}, volume~1, pages 10--13. IEEE, 2004.

\bibitem{simonyan2014very}
Karen Simonyan and Andrew Zisserman.
\newblock Very deep convolutional networks for large-scale image recognition.
\newblock {\em arXiv preprint arXiv:1409.1556}, 2014.

\bibitem{ioffe2015batch}
Sergey Ioffe and Christian Szegedy.
\newblock Batch normalization: Accelerating deep network training by reducing
  internal covariate shift.
\newblock In {\em International Conference on Machine Learning}, pages
  448--456, 2015.

\bibitem{srivastava2014dropout}
Nitish Srivastava, Geoffrey~E Hinton, Alex Krizhevsky, Ilya Sutskever, and
  Ruslan Salakhutdinov.
\newblock Dropout: a simple way to prevent neural networks from overfitting.
\newblock {\em Journal of machine learning research}, 15(1):1929--1958, 2014.

\bibitem{jia2014caffe}
Yangqing Jia, Evan Shelhamer, Jeff Donahue, Sergey Karayev, Jonathan Long, Ross
  Girshick, Sergio Guadarrama, and Trevor Darrell.
\newblock Caffe: Convolutional architecture for fast feature embedding.
\newblock In {\em Proceedings of the 22nd ACM international conference on
  Multimedia}, pages 675--678. ACM, 2014.

\bibitem{duchi2011adaptive}
John Duchi, Elad Hazan, and Yoram Singer.
\newblock Adaptive subgradient methods for online learning and stochastic
  optimization.
\newblock {\em Journal of Machine Learning Research}, 12(Jul):2121--2159, 2011.

\end{thebibliography}
\end{document}